\documentclass[conference]{IEEEtran}
\IEEEoverridecommandlockouts
\usepackage{cite}
\usepackage[colorlinks=true,allcolors={blue}]{hyperref}
\usepackage{amsmath,amssymb,amsfonts}
\usepackage{commath}
\usepackage{algorithmic}
\usepackage{graphicx}
\usepackage{textcomp}
\usepackage{xcolor}
\usepackage{tabularx, booktabs}
\newcolumntype{C}{>{\centering\arraybackslash}X} % centered version of "X" type
\usepackage{url}
\def\BibTeX{{\rm B\kern-.05em{\sc i\kern-.025em b}\kern-.08em
    T\kern-.1667em\lower.7ex\hbox{E}\kern-.125emX}}
\makeatletter
\newcommand{\linebreakand}{%
  \end{@IEEEauthorhalign}
  \hfill\mbox{}\par
  \mbox{}\hfill\begin{@IEEEauthorhalign}
}
\makeatother
\begin{document}

\title{An Evaluation of Model Compression \& Optimization Combinations}
\author{\IEEEauthorblockN{1\textsuperscript{st} Arhum Ishtiaq}
\IEEEauthorblockA{\textit{DSSE, Computer Science} \\
\textit{Habib University}\\
Karachi, Pakistan \\
ai05182@st.habib.edu.pk}
\and
\IEEEauthorblockN{2\textsuperscript{nd} Sara Mahmood}
\IEEEauthorblockA{\textit{DSSE, Computer Science} \\
\textit{Habib University}\\
Karachi, Pakistan \\
sm05155@st.habib.edu.pk}
\and
\IEEEauthorblockN{3\textsuperscript{rd} Maheen Anees}
\IEEEauthorblockA{\textit{DSSE, Computer Science} \\
\textit{Habib University}\\
Karachi, Pakistan \\
ma05156@st.habib.edu.pk}
\and
\IEEEauthorblockN{4\textsuperscript{th} Neha Jafry}
\IEEEauthorblockA{\textit{DSSE, Computer Science} \\
\textit{Habib University}\\
Karachi, Pakistan \\
nj05165@st.habib.edu.pk}
\linebreakand
\IEEEauthorblockN{5\textsuperscript{th} Khubaib Naeem Kasbati}
\IEEEauthorblockA{\textit{DSSE, Computer Science} \\
\textit{Habib University}\\
Karachi, Pakistan \\
kk04333@st.habib.edu.pk}
\and
\IEEEauthorblockN{6\textsuperscript{th} Dr. Abdul Samad}
\IEEEauthorblockA{\textit{DSSE, Computer Science} \\
\textit{Habib University}\\
Karachi, Pakistan \\
abdul.samad@sse.habib.edu.pk}
}

\maketitle

\begin{abstract}
With time, machine learning models have increased in their scope, functionality and size. Consequently, the increased functionality and size of such models requires high-end hardware to both train and provide inference after the fact. This paper aims to explore the possibilities within the domain of model compression, discuss the efficiency of combining various levels of pruning and quantization, while proposing a quality measurement metric to objectively decide which combination is best in terms of minimizing the accuracy delta and maximizing the size reduction factor.
\end{abstract}


\section{Introduction} Deep Neural Networks have been efficiently and increasingly used in various applications of machine learning,  computer vision, and other real-time applications. However, DNNs require high computational power and training and some of those models even on a high end machine might take hours and sometimes days to train. Not only they are intensive in terms of computation but also memory intensive due to a large number of weights taking up considerable storage and memory bandwidth. This makes it difficult to deploy them in resource-constrained environments like embedded systems. To address this limitation, techniques and methodologies for model compression have been attempted to reduce the storage requirement of deep neural networks without impacting the original accuracy.

Model compression is primarily done in two main ways: pruning and quantization. Pruning is the technique that helps develop smaller networks by zeroing out model weights during the training process to achieve model sparsity, eliminating specific connections between neurons. Quantization on the other hand, is the process of mapping values from a large set to a smaller set, thereby decreasing the precision of the information while reducing the model size. Models weights are typically stored as 32-bit floating point numbers and a common approach is to reduce these to either 16-bit floating points or 8-bit fixed points, reducing memory footprint by up to 4 times. 

In this paper, we will focus on experimenting and testing pruning and quantization by utilizing previous works, and comparing their impact on the accuracy of the network models pre- and post-compression. We will also devise a metric that will help objectively measure the best possible combination of pruning and quantization of a model based on the achieved size reduction and consequent accuracy trade-off. 

\section{Related Works}
Several work has been done regarding model compression. Tailin Liang and et al. in 2021 conducted a survey to discuss pruning and quantization methods in \cite{b1}. Tailin Liang et al. provide a comprehensive review of the two model compression techniques including the quantization and  pruning methods used for the experimentation of this paper. The two techniques are described below:
\begin{enumerate}
    \item \textbf{Quantization:} Precision quantization is the reduction of the precision value of the weights of a neural network from the usual FP32 to some lower precision level, often INT8. This conversion to a lower bit representation can enable models to run on smaller devices by reducing bandwidth, energy and on-chip size.
    
    Generally, the clamp function is used to reduce the floating point value during quantization and a zero-point is used to adjust the true zero in the asymmetric mode to cover the whole range.
    \item \textbf{Network Pruning:} One network pruning method discussed in the survey is magnitude based pruning. Magnitude based pruning is a method that identifies the weights that are small enough to not have much impact on the overall evaluation and remove these non essential weights from the evaluation. One simple pruning method is to simply prune zero valued weights or weights that fall under a specified threshold.  
\end{enumerate}
The paper aptly summarises that pruning reduces the number of non-essential computations from the evaluation whereas quantization reduces computation by reducing the complexity of computations due to precision of the datatypes.

Michael. H. Zhu and Suyog Gupta in \cite{b2} discuss that there has been a greater interest in magnitude based weight pruning for network pruning recently owing to the computational efficiency and scalabilty of such techniques. In \cite{b2} Zhu and Gupta provide a comparison between the performance of large-sparse models and small dense model on a large variety of datasets showing that large sparse models have a higher accuracy compared to the former, despite the space overhead. 

In a paper from Kozlov, et al. \cite{b4}, the authors discuss Neural Network Compression Framework (NNCF) for fast model inference. This framework includes the quantization and pruning methods for model compression. In NNCF quantization is represented by affine mapping. Kozlov et al. used two modes of quantization: symmetric and asymmetric modes. The symmetric mode is simpler to implement, since the zero-point is zero, but the asymmetric mode has the advantage of utilizing the complete quantization range and consequently have a better accuracy. For a better trade-off between accuracy and performance, we can apply different levels of precision (e.g. 8, 6, 4 bits) to different layers based on the sensitivity of the layer.

The results of the experimentation using the ImageNet dataset on multiple models showed that INT8 quantization of EfficientNet-B0 model was applied and there was a considerable difference in the accuracy drop between symmetric and asymmetric modes, $0.75\%$ and $0.21\%$ respectively meaning asymmetric quantization has a higher accuracy. 

Song Han et al. (2016) in \cite{b5} introduced a three-stage compression pipeline: pruning, quantization and Huffman coding, termed as 'Deep Compression' to address this limitation by reducing the storage requirement of deep neural networks without impacting the original accuracy. 

For the first stage i.e pruning, Han et al. started by training connectivity, then connections with weights below a threshold were pruned from the network and finally the network was retrained to learn the final weights for the remaining sparse connections. For further compression, they stored the index difference instead of the absolute positions. For the second stage i.e quantization, they further compressed the pruned network by reducing the number of bits, from 32 to 5, required for weight representation. By enforcing weight sharing between multiple connections and then they limited the number of effective weights required to be stored. Finally they applied Huffman coding for encoding weights and indices which further improves the compression rate. 

Result from application on LENET-300-100, LENET-5, AlexNet and VGG-16 showed that the compression pipeline saved 35x to 49x parameter storage across these four networks with no loss of accuracy. Most of the savings comes from pruning and quantization, while Huffman coding gives a marginal gain. On pruning and quantization individually, the accuracy of the pruned network as well as the quantized network drops significantly when compressed below 8\% of the original size. But when combined, the network can be compressed to 3\% of original size without losing accuracy. This showed that quantization and pruning works well together. Further the results were compared with SVD, which is inexpensive but gives very poor compression rate. 

Soheil Hashemi et al. (2016) in \cite{b8} conducted a study to examine the effects of precision quantization on the accuracy of the model as well as the energy saved after quantization. They observed the accuracy and energy saved for a range of precision levels, ranging from the original 32 bit floating point to single bit fixed point. They used three benchmark models for the experiments: using MNIST dataset on LeNet, SVHN dataset on ConvNet and CIFAR-10 dataset on ALEX. The results from the experiment shown in Table IV in the paper show that on reducing the precision even as little as reducing to 32-bit fixed point from floating point gave a significant increase in energy savings, increasing by 12.86\%. A precision level of common interest, 8-bit fixed point, showed energy savings of 85.41\%. Furthermore, the accuracy at some precision level even indicated a slight increase indicating that reducing the precision even down to 8-bit fixed point can have minimal loss in accuracy..

\section{Experiment \& Results}
\begin{figure*}[ht]
    \centering
    \includegraphics[width=\textwidth]{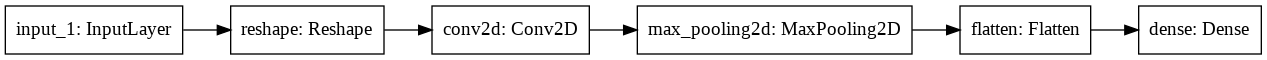}
    \caption{Visual representation of CNN model architecture}
    \label{cnn-vis}
\end{figure*}
\begin{figure*}[ht]
    \centering
    \includegraphics[width=\textwidth]{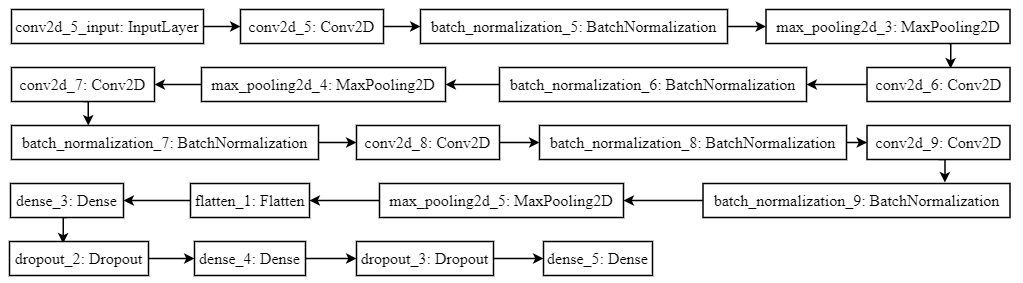}
    \caption{Visual representation of AlexNet model architecture}
    \label{alex-vis}
\end{figure*}
Our main focus is to test the efficacy of pruning and quantization, both individually and in tandem. For the scope of this paper, we will apply the aforementioned techniques on a simple CNN and a state of the art model, AlexNet. For both the models, we first focused on establishing a baseline model with 0$\%$ sparsity and no post-training quantization. We then pruned our models and retrained them, followed by post-training quantization with another round of retraining.

For this experiment we tested out several combinations of hyperparameters, the initial model training was done for a total of 12 epochs with a validation set split being of 30\% of the training dataset. For models that were pruned, the process occurred over 12 epochs with a batch size of 128 and an initial sparsity of 50\%. Each model was then evaluated on test data points in order to calculate the accuracy rate, which was then compared to the test set accuracy of the baseline model for the accuracy delta metric to showcase the impact of each approach on the overall performance of the model. The second metric was the reduction factor which was calculated by tabulating the raw size of gzipped versions of all our trained models and then compared against the baseline model size by calculating the ratio between the sizes. 

Finally, in order to objectively compare the model performance given the various pruning and quantization combinations, we devised a metric that considers the sparsity percentage, weight precision, accuracy delta and the reduction factor to provide a value that can be tangibly compared against each other to decide which combination has the smallest performance trade-off and the most efficiency gain, as shown in Eq \ref{eqn:metric}.

\begin{equation}
\bigtriangleup acc = \textit{accuracy}_n - \textit{accuracy}_{baseline}
\end{equation}
\begin{equation}
r_{n} = \textit{reduction factor of a given model}
\end{equation}
\begin{equation}
s_{n} = \textit{sparsity of a given model}
\end{equation}
\begin{equation}
p_{n} = \textit{weight precision of a given model}
\end{equation}
\begin{equation}\label{eqn:metric}
\textit{Q(x)} = \dfrac{(s_{n} + 8/p_{n})}{2} * tanh(\bigtriangleup acc) * sigmoid(r_{n})
\end{equation}

The main goal with creating a quality metric was to allow comparison not just within different pruning and quantization combinations of a single model but to also ensure standardization across other models. Thus, we divide the formula into two parts, the 'compression' parameter and the 'performance' parameter. The compression parameter is the sum of the sparsity level and the reciprocal of p\_{n} scaled by 8, the sum is then divided by two to scale the parameter between 0 to 1. The performance parameter is a product of the accuracy delta and reduction factor scaled by the tanh and sigmoid function, respectively. The resulting parameters are then multiplied together to give us the final value in the range of -1 to 1. We can then use this quality metric to compare the various combinations of pruning and quantization to find the best model with the most efficient size and accuracy trade-off.  

\subsection{CNN on MNIST}
Fig \ref{cnn-vis} is a visualization of the CNN model architecture, as implemented in our project. We carried out our experiments on the MNIST dataset for handwritten digit classification.
 
Table \ref{tab:cnn-red} show the results observed with different hyperparameters. We can see that the smallest delta is of the 16-bit and the 8-bit quantized model with 0$\%$ sparsity, i.e 0.37 whereas the highest is of the model with 99$\%$ sparsity, leading to a drop of 68.42$\%$ in test set accuracy, in contrast with the baseline model. Highlighted in table \ref{tab:cnn-red}, increasing sparsity and reducing precision causes loss in information representation of our model weights which consequently leads to consistent drops in model performance i.e the increasing negative accuracy delta. 

\begin{table}[ht]
 \caption{Size and Accuracy of CNN upon Pruning and Quantization}
\label{tab:cnn-red}
\begin{center}
\begin{tabular}{cccccc}
\toprule
\multicolumn{1}{p{1cm}}{\centering Model \\ Sparsity } & \multicolumn{1}{p{1cm}}{\centering Precision \\(bits) } & \multicolumn{1}{p{1cm}}{\centering Size \\(bytes) } & \multicolumn{1}{p{1cm}}{\centering Reduction \\Factor }  & \multicolumn{1}{p{1cm}}{\centering Accuracy \\(\%) } & \multicolumn{1}{p{1cm}}{\centering $\Delta$Accuracy \\(\%) } \\ 
\midrule
0   & 32 & 78,170          & - & 98.17 & -  \\
0   & 16      & 17,626  & 4.43 & 97.80 & -0.37 \\ 
0   & 8      & 9,078          & 8.61 & 97.80 & -0.37  \\ 
0.5   & 32      & 35,361          & 2.21 & 97.20 & -0.97 \\
0.5   & 16      & 23,391          & 3.34 & 97.20 & -0.97  \\
0.5   & 8      & 8,173          & 9.56 & 97.15 & -1.02 \\
0.75   & 32      & 29,240          & 2.67 & 97.00 & -1.17 \\
0.75   & 16      & 19,095          & 4.09 & 97.00 & -1.17   \\
0.75   & 8      & 7,563          & 10.34 & 96.99  & -1.18  \\
0.90   & 32      & 17,588          & 4.44 & 95.95 & -2.22   \\
0.90   & 16      & 11,546          & 6.77 & 95.95 & -2.22   \\
0.90   & 8      & 5,331          & 14.66 & 95.95 & -2.22  \\
0.95   & 32      & 12,904          & 6.06  & 84.57 & -13.60  \\
0.95   & 16      & 8,601          & 9.09  & 84.57 & -13.60  \\
0.95   & 8      & 3,851          & 20.30  & 84.57 & -13.60 \\
0.99   & 32      & 8,809          & 8.87 & 29.75& -68.42  \\
0.99   & 16      & 6,270          & 12.47 & 29.75 & -68.42 \\
0.99   & 8      & 2,250          & 34.74 & 29.75 & -68.42 \\
\bottomrule
\end{tabular}
\end{center}
\end{table}

It is evident that compared to an unadulterated model, both pruning and quantization allow for huge increase in model size reduction. In our testing, as we only increased the model sparsity, the size of the model significantly reduces with the highest reduction factor being just above 8 times. Similarly, with quantization, we saw an average reduction factor of 6.42 whereas the accuracy remained comparable with the accuracy of the original model. This signifies that quantization can lead to a significantly smaller model, while maintaining the accuracy. The biggest gains, however, were observed when pruning and quantization were both applied, with our smallest reduction factor being higher than the highest factor observed in the individually pruned or quantized models. Using our proposed quality metric, we see that with the most optimal combination of 90\% sparsity and full-integer quantization, we were able to observe the size of the model decreasing by more than 14 times, from 78,170 bytes to a paltry 5,331 bytes. 

Apart from this, table \ref{tab:alex-qual} shows the quality of the model with different combinations of pruning and quantizations. According to the table, the highest quality of model, based on the quality metric defined by eq \ref{eqn:metric}, is the one with 0\% sparsity and 16 bit quantization. Since the CNN model is small and originally does not have a significant amount of trainable parameters, any loss of information through any form of compression methods results in significant drops in accuracy. 

While the quality metric is not a universal value to decide on the best combination of model compression techniques and hyperparameter combinations, we can still extrapolate some semblance of logic that can help us dictate the right choice when it comes to model compression.

\subsection{AlexNet on CIFAR-10}
We carried out our experiment for AlexNet model on CIFAR-10 dataset. Fig \ref{alex-vis} gives the visualization of our AlexNet model architecture.

The results for different combinations of sparsity and precision levels to measure the performance of the model in terms of the accuracy are listed in Table \ref{tab:alex-red}. It can be seen that for AlexNet, the accuracy delta does not vary significantly nor drops consistently as we increase the sparsity level or reduce the precision of the model, in contrast to our baseline model. Therefore, the performance of the model can be maintained without any huge loss in the accuracy and has the best model quality, as per our metric, at 99\% sparsity with 8-bit precision, reinforcing the fact that quantization and pruning work well together. 

\begin{table}[ht]
 \caption{Size and accuracy of AlexNet model upon Pruning and Quantization}
\label{tab:alex-red}
\begin{center}
\begin{tabular}{cccccc}
\toprule
\multicolumn{1}{p{1cm}}{\centering Model \\ Sparsity } & \multicolumn{1}{p{1cm}}{\centering Precision \\(bits) } & \multicolumn{1}{p{1cm}}{\centering Size \\(bytes) } & \multicolumn{1}{p{1cm}}{\centering Reduction \\Factor }  & \multicolumn{1}{p{1cm}}{\centering Accuracy \\(\%) } & \multicolumn{1}{p{1cm}}{\centering $\Delta$Accuracy \\(\%) } \\ 
\midrule
0   & 32      & 656,296,182          & -  & 82.46  & -  \\ 
0   & 16      & 105,670,635          & 6.21   & 83.50  & 1.04  \\ 
0   & 8      & 49,299,763          & 13.31  & 83.20  & 0.74  \\
0.5   & 32      & 145,224,577          & 4.52  & 85.66  & 3.20   \\ 
0.5   & 16      & 77,622,744          & 8.45   & 88.30  & 5.84   \\ 
0.5   & 8      & 37,665,509          & 17.42    & 86.80  & 4.34     \\ 
0.75   & 32      & 112,778,885          & 5.82 & 85.72  & 3.26     \\ 
0.75   & 16      & 61,816,327          & 10.62   & 88.50  & 6.04     \\ 
0.75   & 8      & 29,841,006          & 21.99 & 87.60  & 5.14   \\ 
0.90   & 32      & 74,410,869          & 8.82 & 84.88  & 2.42   \\ 
0.90   & 16      & 42,674,777          & 15.38  & 87.50  & 5.04     \\ 
0.90   & 8      & 19,667,080          & 33.37  & 85.80  & 3.34      \\ 
0.95   & 32      & 66,897,805          & 9.81  & 84.24  & 1.78    \\ 
0.95   & 16      & 38,945,988          & 16.85 & 85.60  & 3.14        \\ 
0.95   & 8      & 17,386,972          & 37.75     & 85.90  & 3.44      \\
0.99   & 32      & 60,892,461          & 10.78    & 82.27  & -0.19    \\ 
0.99   & 16      & 35,555,186          & 18.46  & 85.60  & 3.14       \\ 
0.99   & 8      & 15,638,735          & 41.97    & 86.70  & 4.24    \\ 
\bottomrule
\end{tabular}
\end{center}
\end{table}

Furthermore, Table \ref{tab:alex-red} confirms the trend we observed for CNN by showing a huge increase in model size reduction for both pruning and quantization. It can be observed that for AlexNet, increasing the sparsity level or reducing the precision individually, gives the highest reduction factor of 10.78 and 13.31, respectively. However, when combined, gives better results without much loss in the accuracy of the model. On the basis of our devised quality metric in Eq. \ref{eqn:metric}, we can see that the optimal combination for AlexNet model turns out to be 99\% sparsity and full-integer quantization, reducing the model size by almost 42 times. We hypothesize that this is because AlexNet is a large network with over 58M parameters, any arbitrary loss of representation through pruning and quantization does not end up significantly impacting the accuracy of the model.

\begin{table}[h]t
 \caption{Model Quality}
\label{tab:alex-qual}
\begin{center}
\begin{tabular}{ccccc}
\toprule
Sparsity (\%)     & Precision (bits) & CNN  & AlexNet \\  
\midrule
0   & 32  &  -   & -    \\ 
\textcolor{blue}{\textbf{0}}   & \textcolor{blue}{\textbf{16}} & \textcolor{blue}{\textbf{-0.0875}}  & 0.1941  \\ 
0   & 8 & -0.1770 & 0.3146   \\ 
0.5   & 32 & -0.2530 & 0.3697    \\ 
0.5   & 16 & -0.3616 & 0.4999  \\ 
0.5   & 8 &  -0.5774 & 0.7497  \\ 
0.75   & 32 & -0.3855  & 0.4971  \\ 
0.75   & 16 & -0.5067 & 0.6250    \\ 
0.75   & 8 & -0.7240  & 0.8749       \\ 
0.90   & 32 & -0.5551  & 0.5659   \\ 
0.90   & 16 & -0.6829  & 0.6999     \\ 
0.90   & 8  & -0.9278    & 0.9476   \\ 
0.95   & 32 & -0.5986  & 0.5668   \\ 
0.95   & 16 & -0.7249  & 0.7223  \\ 
0.95   & 8 & -0.9750    & 0.9730 \\
0.99   & 32 &-0.6199   & -0.1164    \\ 
0.99   & 16 &  -0.7450  & 0.7422  \\ 

\textcolor{red}{\textbf{0.99}}   & \textcolor{red}{\textbf{8}} & -0.9950   & \textcolor{red}{\textbf{0.9946}}  \\ 
\bottomrule
\end{tabular}
\end{center}
\end{table}

\section{Conclusion} 
Through this paper, we conducted a series of experiments aimed at trying various combinations of quantization and pruning on two different models: a CNN on the MNIST dataset and AlexNet on the CIFAR-10 dataset. We also made efforts to mathematically derive a quality measurement metric that helps determine what is the best performing model in terms of how small and accurate it is, in contrast to a baseline model. We see that in some cases, like that of our basic CNN, just using one of our two model compression approaches suffices in producing the model with the highest quality metric in comparison to models that have undergone different combinations of pruning and quantization. In contrast, the best AlexNet instance was the one that had went through the highest pruning and was quantized to the lowest precision, showing us that the best model combination need not be the one with just the largest size reduction or a net positive accuracy delta. 

Pruning and quantization are very effective means of model compression and we see that in reasonable scenarios i.e. where model representation is very dense and the number of trainable parameters is significantly high, both pruning and quantization help not just reduce the overall size and complexity of any given model but also in cases where retraining is done, offer an improvement in model accuracy. Finally, our proposed quality metric has helped us extrapolate the best possible composition of sparsity and precision values, help us objectively evaluate and contrast models that have the best size reduction to accuracy delta trade-off. 

A possible future work could explore the addition and efficacy of other model compression methodologies while also extending our quality metric to factor in those newly considered methods and test the magnitude of their impact on the performance and compression of the model by comparing it against the improvements brought on by pruning and quantization.


\begin{thebibliography}{00}
\bibitem{b1} T. Liang, J. Glossner, L. Wang, S. Shi and X. Zhang, "Pruning and Quantization for Deep Neural Network Acceleration: A Survey", arXiv.org, 2021. [Online]. Available: \url{https://arxiv.org/abs/2101.09671}. [Accessed: 26- Apr- 2021].
\bibitem{b2} M. Zhu and S. Gupta, "To prune, or not to prune: exploring the efficacy of pruning for model compression", arXiv.org, 2021. [Online]. Available: \url{https://arxiv.org/abs/1710.01878}. [Accessed: 26- Apr- 2021].
\bibitem{b4} A. Kozlov, I. Lazarevich, V. Shamporov, N. Lyalyushkin and Y. Gorbachev, "Neural Network Compression Framework for fast model inference", arXiv.org, 2021. [Online]. Available: \url{https://arxiv.org/abs/2002.08679}. [Accessed: 26- Apr- 2021].
\bibitem{b5} S. Han, H. Mao and W. Dally, "Deep Compression: Compressing Deep Neural Networks with Pruning, Trained Quantization and Huffman Coding", arXiv.org, 2021. [Online]. Available: \url{https://arxiv.org/abs/1510.00149}. [Accessed: 26- Apr- 2021].
\bibitem{b6} A. Polino, R. Pascanu and D. Alistarh, "Model compression via distillation and quantization", arXiv.org, 2021. [Online]. Available: \url{https://arxiv.org/abs/1802.05668}. [Accessed: 26- Apr- 2021].
\bibitem{b7} P. Stock, A. Joulin, R. Gribonval, B. Graham and H. Jégou, "And the Bit Goes Down: Revisiting the Quantization of Neural Networks", arXiv.org, 2021. [Online]. Available: \url{https://arxiv.org/abs/1907.05686}. [Accessed: 26- Apr- 2021].
\bibitem{b8}S. Hashemi, N. Anthony, H. Tann, R. Iris Bahar and S. Reda, "Understanding the impact of precision quantization on the accuracy and energy of neural networks", arXiv.org, 2021. [Online]. Available: \url{https://arxiv.org/abs/1612.03940}. [Accessed: 26- Apr- 2021].





\end{thebibliography}
\end{document}